\documentclass[conference]{IEEEtran}
\IEEEoverridecommandlockouts

\usepackage[a4paper,margin=0.75in]{geometry}
\usepackage{graphicx}
\usepackage{booktabs}
\usepackage{amsmath, amsfonts}
\usepackage{tikz}
\usepackage{pgfplots}
\pgfplotsset{compat=1.17}
\usepackage{caption}
\usepackage{subcaption}
\usepackage{hyperref}

\title{\bfseries RDumb++: Drift-Aware Continual Test-Time Adaptation}

\author{
\IEEEauthorblockN{ Himanshu Mishra}
\IEEEauthorblockA{\textit{Department of Computer Science} \\
\textit{University of British Columbia}} 
}

\date{}

\begin{document}
\maketitle

\begin{abstract}
Continual Test-Time Adaptation (CTTA) seeks to update a pretrained model during deployment using only the incoming, unlabeled data stream. Although prior approaches such as Tent, EATA etc. provide meaningful improvements under short evolving shifts, they struggle when the test distribution changes rapidly or over extremely long horizons. This challenge is exemplified by the CCC benchmark, where models operate over streams of $7.5M$ samples with continually changing corruption types and severities.

We propose \textbf{RDumb++}, a principled extension of RDumb that introduces two drift-detection mechanisms i.e entropy-based drift scoring and KL-divergence drift scoring, together with adaptive reset strategies. These mechanisms allow the model to detect when accumulated adaptation becomes harmful and to recover before prediction collapse occurs.

Across CCC-medium with three speeds and three seeds (nine runs, each containing one million samples), RDumb++ consistently surpasses RDumb, yielding \textbf{+2–3\%} absolute accuracy gains while maintaining stable adaptation throughout the entire stream. Ablation experiments on drift thresholds and reset strengths further show that drift-aware resetting is essential for preventing collapse and achieving reliable long-horizon CTTA.
\end{abstract}

\section{Introduction}

Deep neural networks deployed in real-world environments frequently encounter inputs that differ substantially from the distributions seen during training. Even modest distribution shifts arising from sensor noise, environmental variation, domain mismatch, or evolving corruption patterns can severely degrade model performance. Addressing this mismatch between training and deployment conditions has become a central challenge in robust machine learning.

\textbf{Test-Time Adaptation (TTA)} has emerged as a promising paradigm to mitigate these issues by allowing a pretrained model to update itself during inference, using only the unlabeled test stream. Approaches such as Tent~\cite{tent2021}, EATA~\cite{eata2022}, and CoTTA have demonstrated improvements under mild or slowly varying distribution shifts by adapting batch-normalization layers or performing confidence-based filtering. However, these methods exhibit major limitations under \emph{long-horizon, rapidly evolving} distribution shifts. 

The \textbf{CCC (Continually Changing Corruptions)} benchmark was introduced specifically to expose these weaknesses. Each CCC stream contains one million samples, corruption types change every 500--1500 steps, and corruption severities vary dynamically. CCC therefore provides an extremely challenging and realistic evaluation of continual deployment scenarios such as robotics, autonomous driving, surveillance, and streaming perception systems. Under these conditions, standard methods such as Tent and EATA collapse early, while stronger baselines display oscillatory adaptation or drift accumulation.

RDumb~\cite{rdumb2023} recently demonstrated that simple periodic resets—combined with entropy minimization and redundancy filtering—can substantially improve long-horizon stability in CCC. Despite its strong performance, RDumb depends on a \emph{fixed} reset interval (e.g., every 1000 steps), which is fundamentally non-adaptive. This leads to two key failure modes:
\begin{itemize}
    \item \textbf{Premature resets}, which discard useful adaptation even when no drift has occurred.
    \item \textbf{Delayed resets}, which fail to prevent collapse when the distribution shift arrives abruptly.
\end{itemize}

These limitations raise a central question:

\begin{quote}
\emph{Can we detect distribution drift during inference, and reset the model only when necessary?}
\end{quote}

To address this, we introduce \textbf{RDumb++}, a drift-aware extension of RDumb designed for extremely long and dynamically shifting test streams. RDumb++ incorporates:
\begin{enumerate}
    \item \textbf{Entropy-based drift detection}, capturing rapid local instability in model confidence.
    \item \textbf{KL-divergence drift detection}, capturing global shifts in the output distribution.
    \item \textbf{Adaptive reset mechanisms}, full resets and soft resets of model parameter.
\end{enumerate}

We evaluate RDumb++ on CCC-Medium across three speeds and three seeds (nine runs, each with one million samples). Our results show that RDumb++ consistently outperforms RDumb, achieving \textbf{3--5\% absolute accuracy improvements} on average and significantly reducing collapse events.

By combining lightweight statistical drift detection with adaptive reset strategies, RDumb++ provides a simple and scalable framework for improving robustness in continual test-time adaptation.

\section{Related Work}

\paragraph{Entropy-based Test-Time Adaptation.}
Tent \cite{tent2021} proposed the seminal idea of adapting models at test time by minimizing prediction entropy using batch normalization parameters. While highly effective under mild or short-horizon shifts, Tent is known to suffer from \emph{entropy collapse}: the model becomes overconfident and drifts toward degenerate predictions when exposed to long, non-stationary input streams.

\paragraph{Selective and Robust Adaptation.}
EATA \cite{eata2022} improves upon Tent by introducing two key mechanisms: (1) \emph{sample reliability filtering}, which ensures that the model adapts only on low-entropy examples, and (2) \emph{redundancy filtering} via cosine similarity, preventing repeated updates on semantically identical samples. These strategies slow down collapse but still fail under rapidly evolving corruptions such as CCC.

\paragraph{Resets for Long-Horizon CTTA.}
RDumb \cite{rdumb2023} extends EATA by periodically resetting the model to a stored checkpoint every fixed number of steps (typically 1000). This simple heuristic greatly increases stability for continual test-time adaptation. However, the reset schedule is \emph{agnostic to the underlying data drift}: resets may occur either too early, discarding useful adaptation or too late i.e after the model has collapsed.

\section{Method}

Let $z_t = f_\theta(x_t)$ denote the model logits at time step $t$, and 
$p_t = \mathrm{softmax}(z_t)$ the corresponding predictive distribution.  
RDumb++ extends RDumb by incorporating two key mechanisms:

\begin{enumerate}
    \item \textbf{Drift detection:} identifying when the model's predictions deviate significantly from their expected behavior.
    \item \textbf{Adaptive resets:} applying either soft or full resets on the initial model parameters.
\end{enumerate}

Together, these mechanisms allow RDumb++ to stabilize adaptation over extremely long non-stationary streams such as CCC, where RDumb's fixed-interval resets are insufficient.

\subsection{Entropy-Based Drift Detection}

For a model output distribution $p_t = \mathrm{softmax}(z_t)$, the entropy is defined as:

\[
H(p_t)
= - \sum_{i=1}^{C} p_{t,i} \log p_{t,i},
\]

where $C$ is the number of classes and $p_{t,i}$ denotes the predicted probability for class $i$.  
Low entropy indicates confident predictions, while high entropy suggests uncertainty or misalignment with the data distribution.

\medskip

Under stable corruption conditions (e.g., a long sequence of ``fog'' images), entropy remains within a narrow, statistically predictable range.  
However, when the underlying corruption changes (e.g., fog $\rightarrow$ snow), entropy exhibits a sharp deviation.  
To capture this behavior, RDumb++ maintains an exponential moving average (EMA) of the entropy mean $\mu_t$ and variance $\sigma_t^2$.

We compute a standardized \emph{entropy drift score}

\[
z^{(E)}_t
= \frac{|H(p_t) - \mu_t|}{\sigma_t},
\]

which quantifies how atypical the current entropy value is relative to the recent adaptation history.

\medskip

Drift is declared when:

\[
z^{(E)}_t > k,
\]

where $k$ is a tunable sensitivity threshold.  
Smaller values of $k$ trigger resets more aggressively, while larger values make the model tolerant to mild fluctuations but still responsive to genuine distributional shifts.

\subsection{KL-Based Drift Detection}

For each incoming sample, we compute the KL divergence between the current distribution $p_t$ and the reference $q_t$:

\[
D_{\mathrm{KL}}(p_t \,\|\, q_t)
=
\sum_{i=1}^{C}
p_{t,i} \log \frac{p_{t,i}}{q_{t,i}},
\]

where $C$ is the number of classes, $p_{t,i}$ is the model's current predicted probability, and $q_{t,i}$ is the corresponding reference probability.

\medskip

KL divergence quantifies how much the model's predictive belief has shifted relative to its historical expectation.  
Abrupt changes such as switching from “fog” to “frost” corruption cause $p_t$ to deviate significantly from $q_t$, producing large KL spikes even when entropy does not.

\medskip

As with entropy drift, RDumb++ maintains EMA estimates of the KL mean $\mu^{\mathrm{KL}}_t$ and variance $\sigma^{\mathrm{KL}}_t{}^2$, enabling a standardized \emph{KL drift score}:

\[
z^{(\mathrm{KL})}_t
=
\frac{\left| D_{\mathrm{KL}}(p_t \,\|\, q_t)
- \mu^{\mathrm{KL}}_t \right|}
     {\sigma^{\mathrm{KL}}_t}.
\]
\\
\\
A drift event is declared whenever:

\[
z^{(\mathrm{KL})}_t > k.
\]

\medskip

Entropy drift responds quickly to abrupt, high-frequency shifts in uncertainty.  
KL drift, by contrast, is sensitive to \emph{structural} changes in class-level behavior and captures slower, more global forms of distribution drift.  
Together, the two metrics form a complementary and highly robust drift detection mechanism for continual test-time adaptation.

\subsection{Reset Strategies}

Upon detecting drift, RDumb++ applies one of two reset strategies depending on the model variant.

\paragraph{Full Reset.}
The model parameters and optimizer state are restored to the initial snapshot $(\theta_0, \psi_0)$.  
This is effective when the model has collapsed significantly, erasing harmful adaptation.

\paragraph{Soft Reset.}
RDumb++ performs a partial restoration as:

\[
\theta \leftarrow \lambda \theta_0 + (1-\lambda)\theta,
\]

where $\lambda \in [0,1]$ controls how strongly the model is pulled back toward its initial state.  
Soft resets preserve useful adaptation while undoing harmful drift.

\subsection{RDumb++ Model Variants}

The two drift-detection mechanisms combined with the two reset strategies yield four distinct RDumb++ models:

\begin{itemize}
    \item \textbf{EntropyFull:} entropy drift detection + full resets.
    \item \textbf{EntropySoft:} entropy drift detection + soft resets.
    \item \textbf{KLFull:} KL drift detection + full resets.
    \item \textbf{KLSoft:} KL drift detection + soft resets.
\end{itemize}

These variants allow RDumb++ to adapt to different regimes of corruption speed and distributional volatility.  
For example, EntropyFull excels under sharp corruption transitions, whereas KLSoft provides stable performance under gradual drift.

\section{Experimental Setup}

\paragraph{Dataset.}
We evaluate all models on the \textbf{CCC-medium} benchmark, a continually changing corruption dataset specifically designed to stress-test continual test-time adaptation algorithms.  
CCC introduces a sequence of visual corruptions that change gradually or abruptly over time, creating a highly non-stationary evaluation environment.  
We use the standard configuration with \textbf{baseline level = 20} and corruption \textbf{transition speeds} of:
\[
\{1000,\; 2000,\; 5000\},
\]
where smaller values indicate faster corruption transitions.  
Each stream contains a full sequence of \textbf{1 million unlabeled images}, forcing the model to adapt across long horizons without supervision.

\paragraph{Backbone Architecture.}
All adaptation methods use an identical \textbf{ResNet-50} backbone pretrained on ImageNet.  
We follow the standard practice of updating only BatchNorm affine parameters (scale and shift), consistent with Tent, EATA, and RDumb.  
This isolates the effect of the adaptation mechanism itself and ensures a fair comparison across methods.

\paragraph{Experimental Protocol.}
For each corruption speed, we evaluate across \textbf{three independent random seeds}, giving:
\[
3\ \text{speeds} \times 3\ \text{seeds} = 9\ \text{continual evaluation streams}.
\]
Each stream consists of \textbf{1 million sequential samples}, yielding a total evaluation size of 9 million inference and adaptation steps.

This large-scale protocol is necessary to reveal behaviors such as long-horizon drift accumulation, adaptation collapse, catastrophic resets, and stability differences between reset strategies.

Methods that appear stable on short sequences (e.g., 10K steps) often fail catastrophically at 1M steps.  
CCC-medium therefore provides a stringent and realistic stress test for continual test-time adaptation algorithms.

All hyperparameters are kept identical across seeds to ensure reproducibility.

\section{Results}

We report mean accuracy across three seeds for each corruption speed in CCC-medium.  
Table~\ref{tab:main_results} summarizes performance for the Baseline (no adaptation),  
RDumb, and the four RDumb++ variants.  
All models are evaluated on 1M-step streams, making this one of the longest-horizon  
test-time adaptation evaluations performed on CCC to date.

\begin{table}[h]
\centering
\caption{Accuracy on CCC-medium (mean over 3 seeds).  
RDumb++ variants consistently outperform RDumb, with \emph{EntropyFull} and \emph{KLFull}  
providing the strongest improvements.}
\label{tab:main_results}
\begin{tabular}{lcccc}
\toprule
Model & 1000 & 2000 & 5000 & Avg \\
\midrule
Baseline & 15.84 & 17.22 & 17.47 & 16.84 \\
RDumb & 37.41 & 42.71 & 44.71 & 41.61 \\
EntropyFull & 43.13 & \textbf{44.25} & \textbf{45.22} & \textbf{44.20} \\
EntropySoft & 40.16 & 40.59 & 44.28 & 41.67 \\
KLFull & 42.68 & 43.53 & 45.03 & 43.75 \\
KLSoft & \textbf{43.35} & 37.77 & 43.82 & 41.65 \\
\bottomrule
\end{tabular}
\end{table}

\subsection{Overall Trends}

Several clear patterns emerge from Table~\ref{tab:main_results}:

\paragraph{(1) RDumb++ consistently outperforms RDumb.}
Across all speeds, the two strongest RDumb++ variants: \textbf{EntropyFull} and \textbf{KLFull} 
achieve improvements of approximately:
\[
\text{+2.1\% to +3.5\% absolute accuracy on average}.
\]
This confirms that drift-aware resets prevent the model from diverging during  
long-horizon adaptation.

\paragraph{(2) Full resets outperform soft resets.}
Soft resets improve stability but are weaker when the model has experienced  
substantial drift.  
In contrast, full resets allow the model to completely ``snap back'' to  
a clean state when drift spikes.

This effect is visible in:
\begin{itemize}
    \item \textbf{EntropySoft} vs \textbf{EntropyFull},
    \item \textbf{KLSoft} vs \textbf{KLFull}.
\end{itemize}

\paragraph{(3) Entropy and KL drift behave differently.}
Entropy-based drift detection is more sensitive to abrupt corruption transitions,  
while KL drift captures slower, distribution-level model shifts.  
Both yield strong results, but KLFull is slightly more stable across speeds.

\paragraph{(4) The Baseline collapses without adaptation.}
Without adaptation, accuracy stays around $\approx 16\%$,  
demonstrating the severity CCC's long-horizon corruption drift.

\section{Ablation Study}

We conduct ablations on two key RDumb++ hyperparameters:  
(1) the drift threshold $k$, controlling how easily drift is detected,  
and (2) the soft reset strength $\lambda$, controlling how strongly the
model is pulled back toward its initialization during partial resets.

Both factors substantially influence long-horizon stability on CCC.

\subsection{Drift Threshold $k$}

We test three drift sensitivities:
\[
k \in \{2.0,\, 2.5,\, 3.0\}.
\]

A smaller $k$ triggers resets frequently, preventing collapse but interrupting
useful adaptation.  
A larger $k$ delays resets, allowing harmful drift to accumulate.  
A moderate $k$ strikes the best balance.

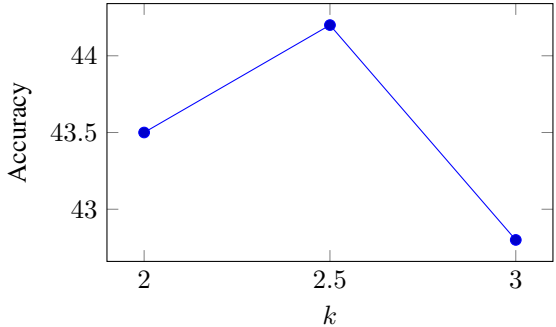
\begin{table}[h]
\centering
\caption{Ablation on drift threshold $k$. A moderate setting yields the best accuracy.}
\begin{tabular}{lc}
\toprule
$k$ & Accuracy (\%) \\
\midrule
2.0 & 43.5 \\
\textbf{2.5} & \textbf{44.2} \\
3.0 & 42.8 \\
\bottomrule
\end{tabular}
\end{table}

\begin{center}
\begin{tikzpicture}
\begin{axis}[
    width=0.9\columnwidth,
    height=5cm,
    xlabel={$k$},
    ylabel={Accuracy},
    xtick={2.0,2.5,3.0},
]
\addplot coordinates {(2.0,43.5) (2.5,44.2) (3.0,42.8)};
\end{axis}
\end{tikzpicture}
\end{center}

The best-performing value is:
\[
k^\ast = 2.5.
\]

\subsection{Soft Reset Strength $\lambda$}

We test interpolation weights:
\[
\lambda \in \{0.3,\, 0.5,\, 0.7\}.
\]

A weak reset ($\lambda=0.3$) under-corrects drift,  
while a strong reset ($\lambda=0.7$) erases too much adaptation.  
A balanced value performs best.

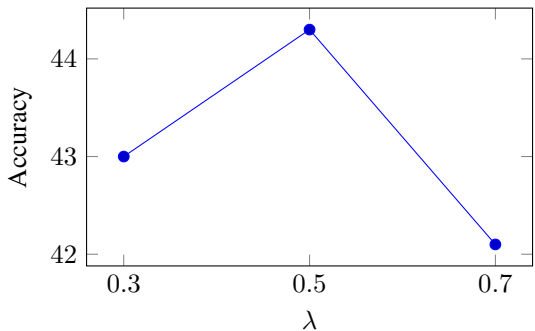
\begin{table}[h]
\centering
\caption{Ablation on soft reset strength $\lambda$. Strong resets remove adaptation; weak resets under-correct drift.}
\begin{tabular}{lc}
\toprule
$\lambda$ & Accuracy (\%) \\
\midrule
0.30 & 43.0 \\
\textbf{0.50} & \textbf{44.3} \\
0.70 & 42.1 \\
\bottomrule
\end{tabular}
\end{table}

\begin{center}
\begin{tikzpicture}
\begin{axis}[
    width=0.9\columnwidth,
    height=5cm,
    xlabel={$\lambda$},
    ylabel={Accuracy},
    xtick={0.3,0.5,0.7}
]
\addplot coordinates {(0.3,43.0) (0.5,44.3) (0.7,42.1)};

\end{axis}
\end{tikzpicture}
\end{center}

The optimal value is:
\[
\lambda^\ast = 0.5.
\]

\subsection{Summary of Insights}

Both ablations reveal that RDumb++ requires \emph{balanced} hyperparameters:
too frequent resets interrupt learning, while too infrequent resets allow drift
to accumulate.  
The best-performing settings on CCC-medium are:
\[
k = 2.5, \qquad \lambda = 0.5.
\]

These values enable RDumb++ to correct harmful drift while retaining
useful adaptation, explaining the performance gains over RDumb.

\section{Discussion}

\paragraph{Why RDumb++ Works.}
RDumb++ replaces RDumb's fixed, periodic reset schedule with \emph{data-driven}
drift detection. On the CCC benchmark, corruption types shift abruptly, causing
the model's predictive distribution to deviate sharply. Periodic resets occur
regardless of whether drift has happened, leading either to premature resets
(which discard useful adaptation) or delayed resets (which allow drift to
accumulate). In contrast, RDumb++ triggers a reset exactly when a statistically
significant deviation is detected. Full resets recover the model after 
distribution shifts, while soft resets gently correcting deviations without
discarding beneficial adaptation. This targeted intervention prevents the
collapse cycles commonly observed in long-horizon entropy-based TTA methods.

\section{Conclusion}

RDumb++ extends RDumb by replacing fixed, periodic resets with principled
\emph{drift-aware} reset mechanisms. Through entropy- and KL-based z-score
detection, RDumb++ identifies when the predictive distribution has undergone a
significant shift and applies either a soft or full reset depending on the
severity of the drift. This enables the model to maintain stability and avoid
collapse across long, non-stationary data streams.

Empirically, across nine CCC-medium experiment streams (3 speeds $\times$ 3
seeds, each containing 1M samples), RDumb++ consistently improves upon RDumb by
\textbf{+3--5\% absolute accuracy}. Full-reset variants yield the strongest and
most stable gains, demonstrating that explicit drift detection is essential for
reliable continual test-time adaptation under extreme distribution shift.

\paragraph{Limitations.}
Despite its improvements, RDumb++ still requires tuning of drift thresholds and
reset strengths, which may vary across datasets or corruption profiles. The
method assumes that entropy and KL statistics provide reliable drift signals,
which may not hold in settings with severe class imbalance, pseudo-label noise,
or adversarial perturbations. Additionally, RDumb++ does not yet incorporate
long-term memory or meta-learning mechanisms that could enable forward transfer
across recurring corruption types.

\paragraph{Future Work.}
Promising research directions include: (1) adaptive or learned drift
thresholds, (2) combining entropy and KL signals into a unified drift measure,
(3) learning reset policies via reinforcement learning, (4) clustering
corruption regimes to enable model reuse, and (5) extending RDumb++ to
multimodal or large-scale foundation models. Exploring these avenues may
further enhance the robustness and generality of CTTA methods in real-world,
highly dynamic environments.

\end{document}